\documentclass[a4paper]{article}

\usepackage{INTERSPEECH2018}

\usepackage{url,hyperref}
\usepackage{soul}
\usepackage{color}
\usepackage{cleveref}

\newcommand{\eg}{e.\,g.,\ }
\newcommand{\ie}{i.\,e.,\ }

\usepackage{tabu,multirow}
\usepackage{arydshln}
\newtheorem{definition}{Definition}

\makeatletter
\def\bstctlcite{\@ifnextchar[{\@bstctlcite}{\@bstctlcite[@auxout]}}
\def\@bstctlcite[#1]#2{\@bsphack
  \@for\@citeb:=#2\do{%
    \edef\@citeb{\expandafter\@firstofone\@citeb}%
    \if@filesw\immediate\write\csname #1\endcsname{\string\citation{\@citeb}}\fi}%
  \@esphack}
\makeatother

\title{Calibrated Prediction Intervals for Neural Network Regressors}
\name{Gil Keren$^1$, Nicholas Cummins$^1$, Bj\"orn Schuller$^{1,2}$}
\address{\fontsize{11}{11}\selectfont
  $^1$ZD.B Chair of Embedded Intelligence for Health Care and Wellbeing, University of Augsburg, Germany\\
  $^2$GLAM -- Group on Language, Audio \& Music, Imperial College London, UK}
\email{cruvadom@gmail.com}

\begin{document}
\bstctlcite{IEEEexample:BSTcontrol}
\maketitle

\begin{abstract}
Ongoing developments in neural network models are continually advancing the state of the art in terms of system accuracy. However, the predicted labels should not be regarded as the only core output; also important is a well-calibrated estimate of the prediction uncertainty. Such estimates and their calibration are critical in many practical applications. Despite their obvious aforementioned advantage in relation to accuracy, contemporary neural networks can, generally, be regarded as poorly calibrated and as such do not produce reliable output probability estimates. Further, while post-processing calibration solutions can be found in the relevant literature, these tend to be for systems performing classification. In this regard, we herein present two novel methods for acquiring calibrated predictions intervals for neural network regressors: empirical calibration and temperature scaling. In experiments using different regression tasks from the audio and computer vision domains, we find that both our proposed methods are indeed capable of producing calibrated prediction intervals for neural network regressors with any desired confidence level, a finding that is consistent across all datasets and neural network architectures we experimented with. In addition, we derive an additional practical recommendation for producing more accurate calibrated prediction intervals. We release the source code implementing our proposed methods for computing calibrated predicted intervals. The code for computing calibrated predicted intervals is publicly available\footnote{Code available in \url{http://github.com/cruvadom/Prediction_Intervals}}.
\end{abstract}

\section{Introduction}
\label{sec:introduction}
Deep learning has undoubtedly improved the state-of-the-art performance of machine learning models across a variety of machine learning applications, in terms of overall system accuracy~\cite{}. In addition, there is an increasing research attention within the deep learning community on estimating prediction uncertainty, \ie recognising and quantifying when an output may be incorrect. The estimation of uncertainty can indeed be crucial for a wide range of applications. For example, the decisions made by neural network technology deployed in healthcare settings could have life-threatening consequences. Uncertainty information could therefore act as a guide for clinicians or doctors to seek a potentially life saving advice.

\begin{figure}[t]
  \centering
  \includegraphics[width=\linewidth]{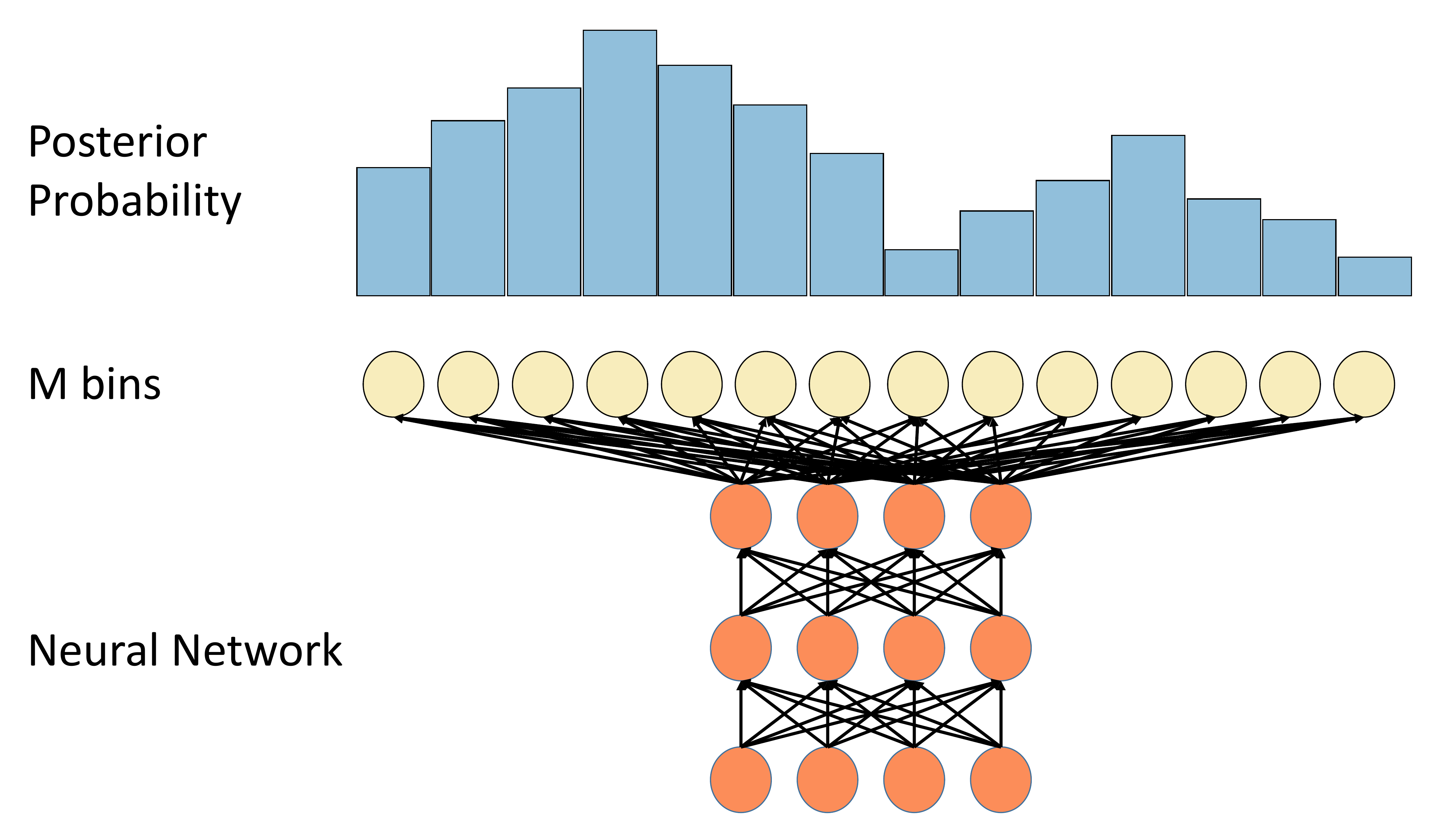}
  \caption{A neural network regressor designed as a softmax classifier. By binning the output space into $M$ bins, one can design a neural network regressor as a softmax classifier over $M$ classes, and derive a posterior distribution over the output space instead of a single point estimate. This allows emitting prediction intervals that contain a prescribed amount of the posterior probability mass. However, we show that the resulting prediction intervals will normally be miscalibrated, \ie will not correspond to the desired confidence level.}
  \label{fig:fig1}
\end{figure}

For a regression problem, uncertainty of a model output can be estimated using \textit{prediction intervals} -- estimates of the interval in which the target label is expected to lie within a prescribed probability. Standard neural network regressors output a point estimation~\cite{DBLP:conf/cvpr/ToshevS14,DBLP:conf/cvpr/HeZRS16,Yao:2018:RRD:3178157.3161181}, from which the estimation of calibrated prediction intervals is a non-trivial task. 
Other neural network regressors use a technique which poses the regression task as a classification task, with a softmax output that produces a posterior distribution over the output space~\cite{pmlr-v48-oord16, vandenOord2016}. Using this method, one could compute prediction intervals for a given confidence level $\alpha$, simply by taking an interval in the output space that contains $\alpha$ of the posterior probability mass, as illustrated in Figure \ref{fig:fig1}. 

However, an interval in the output space that contains $\alpha$ of the posterior probability mass does not have to correspond to a probability of $\alpha$ that the label will fall within this interval's boundaries. For example, a neural network making overconfident predictions may tend to concentrate $\alpha$ of the posterior probability mass in small intervals of the output space, while the probability that these intervals contain the actual labels can be considerably lower. In this case, we say that the prediction intervals are miscalibrated. Recent work has shown that the outputs of modern neural network classifiers are miscalibrated in the sense that posterior class probabilities do not reflect actual correctness probabilities~\cite{DBLP:conf/icml/GuoPSW17}. Therefore, when using neural network regressors that are designed as such classifiers, we expect the resulting prediction intervals to be miscalibrated as well. 

Neural network models have not always been considered miscalibrated. Indeed, work presented in~\cite{DBLP:conf/icml/GuoPSW17, Niculescu05-PGP} identified pre-modern neural network models as a good learning paradigm in terms of producing well-calibrated probabilities for binary classification tasks.
It has been demonstrated that the poor calibration levels observed in more
contemporary deep topologies have come about through recent changes in network architecture and training procedures~\cite{DBLP:conf/icml/GuoPSW17}. For example, miscalibration has been associated with increases in model capacity, and has also been observed in networks trained with batch normalisation or a minimal amount of weight decay~\cite{DBLP:conf/icml/GuoPSW17}.

Despite network calibration being a more recent problem for neural nets, calibration and confidence estimation themselves are not new problems, \eg~\cite{Dawid82-TWB, 596078, platt1999probabilistic, 906002, JIANG2005455, Brummer06-ONO,5749278, Deng12-CMI}. More recently, a plethora of calibration and uncertainty quantification approaches have been proposed and developed for contemporary neural networks in the wider machine learning community. Bayesian Neural Networks produce a probabilistic relationship between the network input and output~\cite{neal2012bayesian, NIPS2017_7141}, but often suffer from tractability issues. Ensemble based approaches, bootstrapping, and Monte Carlo based approaches have also been proposed, for example ~\cite{NIPS2017_7219, 6895153, pmlr-v48-gal16,naumov2017bootstrap}. While such approaches can produce calibrated prediction intervals, they often require training and testing a multitude of different individual networks which considerably increases the associated time and computational costs~\cite{Li2018}. 
Closely related to the current work, a range of post-processing calibration tasks of neural network classifiers were evaluated for a range of different networks topologies~\cite{DBLP:conf/icml/GuoPSW17}. The authors found some of the evaluated methods to successfully calibrate the outputs of the classification models, but counterpart methods for producing calibrated prediction intervals for neural network regressors are still absent. 

Motivated by the above, in this work we present two novel methods for producing calibrated prediction intervals for neural network regressors, at any desired confidence level. Both of our proposed methods are performed as post-processing of the outputs of a the trained regression model that uses a softmax classification layer, therefore do not require retraining of the model, and are very fast to compute. Our first proposed method, \textit{empirical calibration}, assesses the amount of the model's posterior probability mass that corresponds in practice to the desired confidence level. Our second proposed method, \textit{temperature scaling}, is an adaptation of a related method proposed in~\cite{DBLP:conf/icml/GuoPSW17} for calibrating classification models, to the regression and prediction intervals setting. Temperature scaling tunes the smoothness of the model's output distribution, to find a balance that results in calibrated prediction intervals. 

We corroborate our proposed methods in experiments with four regression tasks from the audio and computer vision domains. We first find that as expected, using neural network classifiers to perform regression, and obtaining prediction intervals by taking an interval in the output space that contains the desired posterior probability mass, results in prediction intervals that are poorly calibrated. On the contrary, we find that applying our proposed calibration methods yields prediction intervals that are considerably better calibrated, a finding that is consistent across all datasets, neural network architectures and confidence levels we experimented with. 

Further, we find that when splitting the output space into a finite number of bins, using a larger number of bins and applying our proposed methods results in calibrated prediction intervals that are tighter, i.e., a more accurate estimation of the range in which the label may fall. Finally, we validate that using neural network classifiers to perform regression does not cause any degradation in regression performance, as measured by mean squared error. We conclude that both of are proposed methods are appropriate for emitting calibrated prediction intervals for neural network regressors. We make the source code using empirical calibration and temperature scaling for computing calibrated predicted intervals publicly available\footnote{code available in \url{http://github.com/cruvadom/prediction_intervals}}.

The rest of this paper is laid out as follows. In \Cref{sec:acquiring} we present how regression can be performed using neural network classifiers, and how (not calibrated) prediction intervals can be obtained; \Cref{sec:calibrated} then presents the two proposed calibration methods. The experimental results on the different tasks are presented in \Cref{sec:exp}, and finally a brief conclusion and our future work plans are given in \Cref{sec:conc}.
   
\section{Acquiring Prediction Intervals} \label{sec:acquiring}
\subsection{Posterior Prediction Intervals} \label{sec:posterior}
We consider neural network regressors that process an input $x \in \mathbb{R}^n$ with an associated label $y \in \mathbb{R}$. 
For a regression task, the standard neural network contains a top layer with only one unit ~\cite{DBLP:conf/cvpr/ToshevS14,DBLP:conf/cvpr/HeZRS16,Yao:2018:RRD:3178157.3161181}. The single value in the output of this top layer is then used together with the ground-truth label to compute the mean squared error, which is the training objective of the network. Using this standard design, the network only outputs a single point estimate, and there is no obvious way to use the network's output for computing prediction intervals.

In contrast, a natural approach for designing neural networks regressors from which prediction intervals can be derived, is to construct a regressor that emits a probability distribution $p_x$ over the real numbers, and use this distribution to define intervals with a certain level of probability mass. Denoting $\hat{Y}_x$ as a real-valued random variable that is distributed according to $p_x$, we define the notion of posterior prediction intervals:

\begin{definition} \label{def:posterior} 
The interval $(u_x, v_x)$ is called \textit{posterior $\alpha$-prediction interval} if
\[\mathbb{P}[u(x) < \hat{Y}_x < v(x)] = \alpha \quad \text{and} \quad u(x) < \mathbb{E}[\hat{Y}_x] < v(x).\]
\end{definition}
The posterior prediction interval $(u_x, v_x)$ is simply an interval around the expected prediction of the network $\mathbb{E}[\hat{Y}_x]$ that is designed to contain a probability mass of $\alpha$ from the network's output distribution $p_x$. We refer to $\alpha$ as the \textit{confidence level} of the interval.

With the aim of emitting a probability distribution over the real numbers, neural network regressors can be designed similarly to conventional neural network classifiers, as was done in~\cite{pmlr-v48-oord16,vandenOord2016}. The real numbers are divided into a finite number of bins $M$ with edges 
\[-\infty = a_0 < a_1 < ... < a_M = \infty,\] 
and for the training procedure each real-valued label $y$ is replaced with the appropriate class label $t \in \{0, ... , M - 1\}$ such that 
\[a_t \leq y < a_{t+1}.\] 
The single unit top layer of the standard neural network regressors is replaced with a layer of $M-1$ units. Softmax normalisation is then applied on the output of the top layer and the network is trained as a standard neural network classifier with the cross-entropy loss. The output of such a neural network is a vector of class probabilities 
\[(pr_0, ... , pr_{M-1}).\]
To emit a probability distribution over the real numbers, we can distribute each class probability uniformly, or according to the distribution of training set real-valued labels, between the class's bin boundaries.

Posing the regression problem as a classification problem allows the network to emit a distribution over the real numbers instead of a point estimate, which in turn can be used to calculate posterior $\alpha$-prediction intervals.

\subsection{Calibrated Prediction Intervals} \label{sec:calibrated}

In Section \ref{sec:posterior} we described how neural network regressors can be designed in a manner that allows emitting posterior $\alpha$-prediction intervals. However, the confidence level $\alpha$ does not guarantee that the label is likely to fall within its appropriate posterior prediction interval with probability $\alpha$. For example, consider the case of a neural network that produces overly confident predictions. In this case, the output probability distribution $p_x$ will have most of its mass concentrated in a small region, therefore prediction intervals containing a mass of $\alpha$ of the network's output probability distribution will be very narrow. However, despite the confident predictions, the actual ground-truth labels might fall within the boundaries of those intervals on average only $\alpha_0$ of the times, with $\alpha_0 < \alpha$. Equivalently, the confidence level $\alpha$ may also not represent the actual probability of the label falling within the prediction interval's boundaries in the case of network predictions that are not confident enough.

Ideally, one would aspire to obtain prediction intervals with a confidence level of $\alpha$, such that $\alpha$ is the actual probability of the label falling within the prediction interval's boundaries. We define the notion of calibrated prediction intervals:

\begin{definition} \label{def:calibrated}
A set of intervals $\{(u_x, v_x)\}_{x \in X}$ is considered as \textit{calibrated $\alpha$-prediction intervals} if 
\[\mathbb{P}_{x,y \sim X,Y} [u_x < y < v_x] = \alpha,\]
where $X,Y$ corresponds to the joint distribution of inputs and labels of the given regression task.
\end{definition}

We refer to $\alpha$ as the \textit{confidence level} of the calibrated prediction intervals. In regression analysis, a calibrated prediction interval is an estimate of an interval in which the label will lie, with a certain probability $\alpha$. Calibrated prediction intervals capture information about the uncertainty of the predicted value across the output space, and convey information that is absent from a single point estimate of the label, that might be critical for a wide range of practical applications. 

In recent work, it was shown that modern neural network \textit{classifiers} tend to produce non-calibrated outputs, \ie the posterior probability assigned to predictions does not correspond to the actual ground-truth accuracy of these predictions~\cite{DBLP:conf/icml/GuoPSW17}. Therefore, when using neural network regressors that are constructed as classifiers, and using those to emit posterior $\alpha$-prediction intervals, we cannot expect those posterior $\alpha$-prediction intervals to be calibrated $\alpha$-prediction intervals. In Section \ref{sec:exp} we show that indeed in practice, the obtained posterior $\alpha$-prediction intervals are not calibrated $\alpha$-prediction intervals.

Below we present the main novel contribution of this work, two methods for computing calibrated $\alpha$-prediction intervals for neural network regressors. Consider the neural network regressors designed as classifiers described in Section \ref{sec:posterior}. Recall that in this setting we divide the real numbers into $M$ bins, and given an input $x$, a regressor emits a categorical probability distribution over the different bins: $(pr_0, ..., pr_{M-1})$. 

We compute the network's real-valued prediction (point estimate) as the expected prediction with respect to the emitted class probability distribution:
\begin{equation} \label{eq:yhat}
\hat{y} = \sum \limits _{i=0}^{M-1} pr_i * c_i,
\end{equation}
where $c_i$ is the mean of real-valued labels of all training examples with class label $i$. We denote the class that contains $\hat{y}$ (according to its bin's edges) with $\hat{t}$:
\begin{equation}
\hat{t} = r \quad \text{s.t.} \quad a_{r} \leq \hat{y} \leq a_{r+1}.
\end{equation}

For computing posterior $\alpha$-prediction intervals, we take the smallest symmetric interval around $\hat{t}$ that contains $\alpha$ of the neural network's posterior probability mass. Formally, we take the posterior $\alpha$-prediction interval to be
\begin{equation} \label{eq:pi3}
(u_x^\alpha, v_x^\alpha) = (a_{\hat{t}-i}, a_{\hat{t}+1+i}),
\end{equation}
such that $i$ is the minimal non-negative integer (possibly zero) for which
\begin{equation}
pr_{\hat{t}-i} + ... + pr_{\hat{t}+i} \geq \alpha.
\end{equation}
Note that we restrict the endpoints of the interval to be the discrete bins edges, therefore the condition from Definition \ref{def:posterior} only holds approximately.

In the rest of this section we describe our two proposed novel methods for calibrating the prediction intervals. Both methods apply post-processing to the outputs of a trained neural network, and do not require retraining the neural network. The hyperparameters of the methods are to be chosen using a validation set, and the chosen values should then be used when applying the methods to the test set predictions. 

\subsubsection{Empirical Calibration} We first observe that posterior $\alpha_0$-prediction intervals $(u_x^{\alpha_0}, v_x^{\alpha_0})$ as defined according to (\ref{eq:pi3}) are actually calibrated $\alpha_1$-prediction intervals for
\begin{equation}
\alpha_1 = \mathbb{P}_{x,y \sim X,Y} [u_x^{\alpha_0} < y < v_x^{\alpha_0}].
\end{equation}
This holds because for every set of prediction intervals, there is an actual probability of the label falling within the boundaries of those intervals. Therefore by definition those are calibrated prediction intervals with this probability as their confidence level.

When calibrating the prediction intervals empirically, we want to find $\alpha_0$ such that the posterior $\alpha_0$-prediction intervals are calibrated $\alpha$-prediction intervals, for a desired confidence level $\alpha$. Note that $\mathbb{P}_{x,y \sim X,Y} [u_x^{\alpha_0} < y < v_x^{\alpha_0}]$ is increasing in $\alpha_0$ with fixed points in 0 and 1, since larger posterior prediction intervals necessarily mean that the label is more likely to fall within the intervals' boundaries. 

Therefore, our empirical calibration method is comprised of a binary search along different values of $\alpha_0 \in [0,1]$ to find $\alpha_0$ such that $|\mathbb{P}_{x,y \sim X,Y} [u_x^{\alpha_0} < y < v_x^{\alpha_0}] - \alpha| < \epsilon$ on the validation set, for a given error tolerance $\epsilon$. In our experiments we use $\epsilon=0.001$. The error tolerance is necessary, since for a finite validation set finding calibrated prediction intervals with confidence level \textit{exactly} $\alpha$ may be impossible. The $\alpha_0$ that we end up with is the one that is used for computing prediction intervals on for the test set. 

\subsubsection{Temperature Scaling} When training the neural network for the classification task, class probabilities $(pr_0, ... , pr_{M-1})$ are computed from the output of the top layer $(z_0, ... , z_{M-1})$ using the softmax function:
\begin{equation} \label{eq:softmax1}
pr_i = \frac{\exp(z_i / T)}{\sum \limits _{j=0} ^{M-1} \exp(z_j / T)},
\end{equation}
where $T$ is called the softmax temperature. During training, the default temperature $T=1$ is used. 
Equation \ref{eq:softmax1} can be written as
\begin{equation}
pr_i = \frac{1}{\sum \limits _{j=0} ^{M-1} \exp((z_j - z_i) / T)},
\end{equation}
that shows that the output of the softmax normalisation depends only on the the temperature $T$ and the differences between the output values $(z_0, ... , z_{M-1})$. Therefore, scaling the outputs of the top layer before applying the softmax function affects the smoothness of the output probability distribution. Specifically, using a lower temperature $0 < T < 1$ makes the probability distribution ``pointier'', \ie more probability mass is given to the classes with higher $z$ values. On the contrary, using a larger temperature $1 < T < \infty$ tends towards distributing the probability mass more evenly between the different classes. 

Using this property of the softmax normalisation function, temperature scaling uses a different temperature at evaluation time for computing class probabilities. A network that produces overconfident predictions, will result in posterior $\alpha$-prediction intervals that are too narrow, \ie $\mathbb{P}_{x,y \sim X,Y} [u_x^\alpha < y < v_x^\alpha] < \alpha$.
In this case, temperature scaling with a temperature $T > 1$ can be applied to reduce the network's confidence, and increase the width of the posterior prediction intervals. Equivalently, a low temperature $0 < T < 1$ should be used to increase the network's confidence and decrease the width of posterior prediction intervals. 

More generally, we define 
\begin{equation}
F_\alpha(T) = \mathbb{P}_{x,y \sim X,Y} [u_x^\alpha < y < v_x^\alpha]
\end{equation}
where $u_x^\alpha$ and $v_x^\alpha$ are the posterior $\alpha$-prediction intervals that now depend also on $T$. As increase in $T$ increases the width of the posterior prediction intervals, the function $F_\alpha(T)$ is continuous and monotonic increasing in $T$, with $\lim _{T \rightarrow 0} F_{\alpha}(T) = 0$ and $\lim _{T \rightarrow \infty} F_{\alpha}(T) = 1$.
Therefore, there must be a temperature $T$ such that $F_\alpha(T) = \alpha$.

Motivated by the above theoretical properties, our temperature scaling method is comprised of a binary search along different values of $T$ to find the temperature value such that
\begin{equation}
|F_\alpha(T) - \alpha| < \epsilon
\end{equation}
on the validation set, for the desired confidence level $\alpha$ and a given error tolerance of $\epsilon$. In our experiments we use an error tolerance $\epsilon=0.001$ that is again necessary, since for a finite validation set finding calibrated prediction intervals with confidence level exactly $\alpha$ may be impossible. The temperature $T$ that is chosen using the validation set is then used for computing prediction intervals for the test set. Temperature scaling was used in ~\cite{DBLP:conf/icml/GuoPSW17} for calibrating the output probabilities of neural network \textit{classifiers}, and here we extend this method to the regression and prediction intervals setting.

\section{Experiments} \label{sec:exp}
We evaluated our two proposed calibration methods for prediction intervals on four different regression tasks from the audio and computer vision domains. 

\subsection{Datasets and Tasks}
We describe the four regression tasks and datasets we used in our experiments. 

\subsubsection{Age Prediction (Audio)}
The first task we consider is the prediction speakers' age based on a recording of their speech, using the aGender corpus~\cite{burkhardt2010database,Schuller10-TI2}. The aGender corpus contains audio recordings of predefined utterances and natural speech, annotated for the speakers' age and gender. We split the corpus into speaker independent training, validation and test sets, according to the split used in~\cite{DBLP:conf/ijcnn/KerenS16}. In total, the three sets contain more than 38 hours of audio, in more than 53,000 utterances. The total number of speakers is 611, such that 331 speakers are assigned to the training set, 140 to the validation set, and 299 to the test set. We extracted Mel-Frequency Cepstrum Coefficients (MFCC) features from each recordings, using frames of 25\,ms shifted by 10\,ms. From every frame 13 features were extracted. We applied mean and standard deviation normalisation across features and time, for every recording separately. 

\subsubsection{SNR Prediction}
The second regression task from the audio domain we experimented with is prediction of Signal-to-Noise Ratio (SNR) of speech audio utterances with background noise. For constructing this task's corpus, we used clean speech utterances from the degree of nativeness corpus from the INTERSPEECH 2016 computational paralinguistics challenge~\cite{Schuller16-TI2,keren2016convolutional} and background noise recordings from the CHiME-4 challenge~\cite{DBLP:journals/csl/VincentWNBM17}. The native language corpus contains more than 64 hours of clean speech utterances from 5,132 speakers of 11 different native languages, split into speaker independent training, validation, and test sets. The background noises are recordings of four different environments: bus, caf{\'e}, pedestrian area, street junction, and are 14 hours in total. For creating the training set, training speech utterances were mixed with random segments of the background noises according to a random SNR in the range $[0,25]$. The SNR was then used as the real-valued label for the regression task. The validation and test set were created in a similar manner, using the corresponding clean utterances from the native language corpus and dedicated portions of the noise recordings. We applied a short-time Fourier transform (STFT) on every recording to extract 201 magnitude spectrogram features from every 25\,ms frame, where frames are shifted 10\,ms. The magnitude spectrogram features were then normalised across features and time, for every utterance separately, to have a mean of 0 and a standard deviation of 1. 

\subsubsection{Age Prediction (Images)} \label{sec:ageimages}
The first dataset we experimented with in the computer vision domain is the Wikipedia faces dataset \cite{Rothe-ICCVW-2015}. The dataset contains 62,359 images of people (one image per person) crawled from Wikipedia, labelled with the age of each person at the time the picture was taken. Since the dataset has no official training/validation/test split, we randomly allocated 60\% of the examples to the training set, 20\% to the validation set and 20\% to the test set. As the dimensions of the different images vary, we resized every image to $224 \times 224$ pixels before feeding it to the neural network. In addition, we normalised pixel values for every image separately, to have a mean of 0 and a standard deviation of 1. 

\subsubsection{ISO Speed Prediction}
The second images dataset we experimented with is the MIRFLICKR-25000 dataset. The MIRFLICKR-25000 dataset consists of 25,000 images downloaded from the social photography site Flickr through its public API \cite{huiskes08}. In addition to images, the dataset contains additional metadata on every image, such as the ISO speed, that measures the sensitivity of the camera's film or sensor to light. The ISO speed affects the brightness of photos, therefore a regression task for predicting the ISO speed of given images is sensible. We split the dataset and extracted features in the same way as described in Section \ref{sec:ageimages}.

\subsection{Neural Networks} As described in Section \ref{sec:acquiring}, we learn the regression tasks using a classification neural network, where the real numbers are split into $M$ bins. 
For the audio experiments, the network we used is comprised of two long short-term memory (LSTM) layers, each with 512 units. The output of the last time step in the top layer is fed into the fully-connected output layer, with the number of units equal to the number of bins we use. Softmax normalisation is applied to the output layer's units.

For the computer vision experiments, we used a convolutional neural net (CNN) that is comprised of 8 residual blocks \cite{DBLP:conf/cvpr/HeZRS16}. Each residual block first applies a convolutional layer on the input, followed by batch normalisation \cite{DBLP:conf/icml/IoffeS15} and a rectified linear activation function. A second convolutional layer is then applied on the output of the rectified linear activation, and the output is added to the block's input. Batch normalisation and another rectified linear activation are then applied, to emit the output of the residual block. Before applying the residual blocks, a convolutional layer with a $7 \times 7$ kernel is applied on the network's input, with a $2 \times 2$ stride and 64 feature maps. The output of this convolutional layer is fed to the a sequence of 8 residual blocks, all using convolutional kernel size of $3 \times 3$ and 64,64,128,128,256,256,512,512 feature maps (one value for each residual block). A $2 \times 2$ stride is applied for residual blocks number 3, 5 and 7. A global average pooling is applied on the output of the last residual block, to average each of the 512 feature maps across all spatial locations. Similarly to the audio experiments, a fully-connected layer is then applied to project the 512 dimensional vector to the relevant number of bins, and a softmax normalisation is applied. 

In all experiments, the training objective is the standard cross-entropy, and model parameters are learnt using the Adam optimiser~\cite{kingma2015adamICLR} with default $\beta_1, \beta_2$ values and a learning rate of 0.001. We experimented with binning the real numbers into $M=10,30,60$ bins to demonstrate that our method can operate successfully regardless of the number of bins, and to study the differences between the resulting prediction intervals with different number of bins. For a given number of classes $M$, we set class boundaries $a_0, ... , a_M$ to be equally spaced between the minimum and maximum real-valued label values in the training set, and then set $a_0 = -\infty$ and $a_M = \infty$.

\begin{table}[th!]
\caption{A comparison of test set calibration error ([\%]) before (`Posterior' column) and after applying each of the the two proposed calibration methods for the different regression tasks. `Empirical', `Temp' and `Conf' columns represent empirical calibration, temperature scaling and the prediction intervals' confidence level respectively.
In all cases, both of the proposed methods manage to considerably reduce the calibration error of prediction intervals, compared to prediction intervals based on the networks' posterior distribution (smaller numbers on the right side of the dashed line). Both of the proposed methods yield comparable performance. This result holds when training the network with either 10, 30 or 60 bins, with no clear advantage for a specific number of bins.
}
\label{tab:tab1}
  \centering
  \begin{tabu}{|c|c|c|[1.2pt]c:c|c|}
\hline
Dataset & Conf' & Bins & Posterior & Empirical & Temp' \\
\hline
\multirow{9}{*}{\begin{tabular}{@{}c@{}}Age \\ Audio \end{tabular}} 
& \multirow{3}{*}{66\%} 
& 10 & 7.63 & 0.60 & 0.09 \\
& & 30 & 12.40 & 2.25 & 1.80 \\
& & 60 & 11.95 & 0.82 & 0.35 \\
\cline{2-6}
& \multirow{3}{*}{80\%} 
& 10 & 10.78 & 0.64 & 1.16 \\
& & 30 & 15.37 & 2.63 & 3.13 \\
& & 60 & 13.74 & 1.45 & 2.69\\
\cline{2-6}
& \multirow{3}{*}{90\%} 
& 10 & 9.64 & 1.81 & 2.44  \\
& & 30 & 11.83 & 2.53 & 3.13 \\
& & 60 & 10.97 & 1.95 & 2.16\\
 \hline
 
\multirow{9}{*}{SNR}
& \multirow{3}{*}{66\%} 
& 10 & 22.11 & 6.44 & 7.22 \\
& & 30 & 19.67 & 1.78 & 1.78 \\
& & 60 & 12.22 & 0.11 & 0.78 \\
\cline{2-6}
& \multirow{3}{*}{80\%} 
& 10 & 14.56 & 1.67 & 0.56 \\
& & 30 & 12.67 & 2.89 & 1.78 \\
& & 60 & 9.22 & 1.44 & 1.78 \\
\cline{2-6}
& \multirow{3}{*}{90\%} 
& 10 & 7.56 & 0.89 & 0.44 \\
& & 30 & 6.11 & 2.78 & 2.78 \\
& & 60 & 4.78 & 0.00 & 0.11 \\
 \hline
 
\multirow{9}{*}{\begin{tabular}{@{}c@{}}Age \\ Images \end{tabular}} 
& \multirow{3}{*}{66\%} 
& 10 & 0.29 & 0.01 & 0.14 \\
& & 30 & 4.50 & 0.14 & 0.26 \\
& & 60 & 13.43 & 0.22 & 0.29 \\
\cline{2-6}
& \multirow{3}{*}{80\%} 
& 10 & 3.90 & 0.15 & 0.05 \\
& & 30 & 2.71 & 0.26 & 0.14 \\
& & 60 & 14.98 & 0.22 & 0.63 \\
\cline{2-6}
& \multirow{3}{*}{90\%} 
& 10 & 4.46 & 0.11 & 0.08 \\
& & 30 & 0.73 & 0.08 & 0.05 \\
& & 60 & 14.34 & 0.25 & 0.02 \\
 \hline
 
\multirow{9}{*}{\begin{tabular}{@{}c@{}}Iso \\ Speed \end{tabular}} 
& \multirow{3}{*}{66\%} 
& 10 & 17.39 & 0.76 & 0.82 \\
& & 30 & 6.06 & 0.06 & 0.70 \\
& & 60 & 7.21 & 0.21 & 0.58 \\
\cline{2-6}
& \multirow{3}{*}{80\%} 
& 10 & 6.58 & 0.15 & 0.67 \\
& & 30 & 6.58 & 0.15 & 0.67 \\
& & 60 & 4.45 & 0.03 & 0.06 \\
\cline{2-6}
& \multirow{3}{*}{90\%} 
& 10 & 3.55 & 0.21 & 0.24 \\
& & 30 & 4.06 & 0.73 & 0.27 \\
& & 60 & 3.70 & 0.24 & 0.91 \\
 \hline
 		
  \end{tabu}
\end{table}

\subsection{Calibration Results}
For the main results of this work, we evaluated each of the two proposed calibration methods from Section \ref{sec:calibrated} on the different regression tasks, with different neural network architectures and different number of bins. For each task, we trained three neural networks with 10, 30 and 60 bins. Each of the proposed calibration methods was applied to the outputs of each trained network using confidence levels of 66\%, 80\% and 90\%. For each calibration method and dataset, the associated hyperparameters were chosen using the validation set, then we applied this calibration method to the test set using the chosen hyperparameters. All results we report are on the test set. 

The aim of each calibration method is to produce calibrated $\alpha$-prediction intervals. To assess the level in which this goal was achieved, we measure the \textit{calibration error}, which is the absolute difference between the desired confidence level $\alpha$ and the actual probability of the label falling within the boundaries of the acquired prediction intervals. Mathematically, the calibration error is defined as
\begin{equation}
| \mathbb{P}_{x,y \sim X,Y} [u(x) < y < v(x)] - \alpha |,
\end{equation}
where $(u(x), v(x))$ is the prediction interval emitted by the calibration method for example $x$, and $X,Y$ are distributed uniformly over the test set examples. 

A comparison of the calibration error when using the posterior prediction intervals, and after applying each of the two proposed calibration methods is given in Table \ref{tab:tab1}. First, we observe that the posterior prediction intervals, without applying a calibration method, generally yield a large calibration error. This finding is consistent with findings from~\cite{DBLP:conf/icml/GuoPSW17} regarding the miscalibration of modern neural network classifiers. Second, we see that in all cases, both the empirical calibration and temperature scaling methods manage to considerably reduce the calibration error, eliminating the calibration error to small levels of normally around 0\%-2\%. These results indicate that using these methods, calibrated prediction intervals for neural network regressors can indeed be acquired. Moreover these findings hold across all datasets, confidence levels, and number of bins used for training the networks. 

However, even when using one of the two proposed calibration methods, calibration error does not vanish completely. The reason for this is that calibration hyperparameters were chosen on the validation set, and do not generalise perfectly to the test set. Nevertheless, a calibration error of 1\%-2\% is sufficiently enough for the majority of applications (e.g., a confidence level of 81\% instead of a desired 80\% will not make a large difference in most applications). Both calibration methods yield comparable performance, and are fast to execute, typically around 1-3 seconds for a test set of 10000 examples, depending on the number of bins used. 

\begin{table}[th!]
  \caption{A comparison of the test set average width of prediction intervals using the two proposed calibration methods, empirical calibration and temperature scaling. `Empirical', `Temperature' and `Confidence' columns represent empirical calibration, temperature scaling and the prediction intervals' confidence level respectively. For all datasets except `Age Audio', training the network with more bins generally results in tighter prediction intervals, since the network can learn a more precise distribution of posterior probability (numbers in the 30 and 60 bins rows are generally smaller than in the 10 bins rows). The width of the intervals is comparable between the two calibration methods and naturally grows with the confidence level. Finally, the width of the intervals naturally depends on the performance of the neural network in the regression task.  
 }
  \label{tab:tab2}
  \centering
  \begin{tabu}{|l|c|c|[1.2pt]c|c|}
\hline
Dataset & Confidence & Bins & Empirical & Temperature \\
\hline
\multirow{9}{*}{\begin{tabular}{@{}c@{}}Age \\ Audio \end{tabular}} 
& \multirow{3}{*}{66\%} 
& 10 & 34.84 & 35.70 \\
& & 30 & 34.05 & 36.35 \\
& & 60 & 36.16 & 38.59 \\
\cline{2-5}
& \multirow{3}{*}{80\%} 
& 10 & 44.55 & 44.79 \\
& & 30 & 43.84 & 44.13 \\
& & 60 & 45.23 & 45.41 \\
\cline{2-5}
& \multirow{3}{*}{90\%} 
& 10 & 52.77 & 52.68 \\
& & 30 & 52.69 & 52.30 \\
& & 60 & 53.84 & 54.32 \\
 \hline
 
\multirow{9}{*}{SNR} 
& \multirow{3}{*}{66\%} 
& 10 & 2.60 & 2.60 \\
& & 30 & 1.97 & 1.91 \\
& & 60 & 1.64 & 1.58 \\
\cline{2-5}
& \multirow{3}{*}{80\%} 
& 10 & 3.49 & 3.31 \\
& & 30 & 2.50 & 2.47 \\
& & 60 & 2.22 & 2.15 \\
\cline{2-5}
& \multirow{3}{*}{90\%} 
& 10 & 4.74 & 4.63 \\
& & 30 & 3.11 & 3.04 \\
& & 60 & 2.96 & 2.94 \\
 \hline
 
\multirow{9}{*}{\begin{tabular}{@{}c@{}}Age \\ Images \end{tabular}}
& \multirow{3}{*}{66\%} 
& 10 & 20.99 & 20.92 \\
& & 30 & 19.66 & 19.11 \\
& & 60 & 18.71 & 19.80 \\
\cline{2-5}
& \multirow{3}{*}{80\%} 
& 10 & 27.48 & 27.65 \\
& & 30 & 28.83 & 25.53 \\
& & 60 & 25.71 & 25.81 \\
\cline{2-5}
& \multirow{3}{*}{90\%} 
& 10 & 35.60 & 35.43 \\
& & 30 & 33.51 & 33.58 \\
& & 60 & 34.63 & 33.72 \\
\hline

\multirow{9}{*}{\begin{tabular}{@{}c@{}}Iso \\ Speed \end{tabular}} 
& \multirow{3}{*}{66\%} 
& 10 & 2.23 & 2.20 \\
& & 30 & 1.94 & 1.80 \\
& & 60 & 2.21 & 2.08 \\
\cline{2-5}
& \multirow{3}{*}{80\%} 
& 10 & 3.22 & 3.23 \\
& & 30 & 2.94 & 3.02 \\
& & 60 & 2.90 & 2.91 \\
\cline{2-5}
& \multirow{3}{*}{90\%} 
& 10 & 4.35 & 4.44 \\
& & 30 & 4.09 & 4.05 \\
& & 60 & 3.83 & 3.79 \\
\hline
 		
\end{tabu}
\end{table}

Further, we compare the width of the emitted prediction intervals for the empirical calibration and temperature scaling methods. Table \ref{tab:tab2} contains the average width of the prediction intervals for test sets of the different regression tasks. 
Posterior prediction intervals were above to be poorly calibrated, therefore their width is not meaningful with respect to the desired confidence level, and we omit them from Table \ref{tab:tab2}.
We first observe that naturally, the width of the interval grows with the desired confidence level. The main conclusion that can be derived from these results is that networks trained using a larger number of bins tend to produce tighter prediction intervals. Specifically, for all tasks except age prediction from audio signal, the width of the resulting calibrated prediction intervals is generally smaller when using 30 or 60 bins, compared to 10 bins. The reason for this phenomenon is that a larger number of bins allows the network a more precise allocation of posterior probability mass.

Additionally, we find that the two calibration methods produce prediction intervals of a comparable width, with no prominent advantage for neither of the two methods. This result indicates that both methods can be interchangeably used to produce calibrated prediction intervals of the same quality. Lastly, we note that width of the prediction intervals is closely affected by the quality of the regressor that they are based on. A better neural network regressor is one that assigns a higher probability mass around the correct labels, which will in turn result in tighter prediction intervals.

\subsection{Regression Results}
For studying the the effect of performing the regression tasks using neural network classifiers, we additionally train a standard neural network regressor for each of the tasks. For each task the standard neural network regressor is trained with an identical architecture to the corresponding neural network classifier for this task, except the topmost layer that contains only a single unit, as described in Section \ref{sec:acquiring}. The regressor is trained with the same optimiser as the classifiers to minimise the mean squared error (MSE) between the network's predictions and the labels. For the classification models, MSE is computed using the prediction $\hat{y}$ defined in Eq. \ref{eq:yhat}. 

The root MSE on the test set for the different models is found in Table \ref{tab:tab3}. The results in the table show that regression performance of the standard regressor and the classifiers is generally comparable on all tasks. We therefore conclude that training neural network regressors using neural network classifiers, that allow emitting calibrated prediction intervals, does not cause any degradation in the regression task performance.

\begin{table}[th!]
\caption{Performance in the different regression tasks as measured by the root MSE, for a standard neural network regressor and a neural network classifier with different number of classes (denoted as cls'). The performance of the standard regressors is comparable to the performance of the models performing regression using a classification models. This indicates that using neural network classifiers to perform regression task, that allow emitting calibrated prediction intervals, does not cause any degradation in the regression performance.}
  \label{tab:tab3}
  \centering
  \begin{tabular}{|l|r|r|r|r|}
    \hline
    Dataset & Standard & 10 cls' & 30 cls' & 60 cls'\\
    \hline
    Age (Audio) 	& 20.07 & 19.73 & 19.95 & 20.04 \\
    SNR	     		& 1.32 & 1.21 & 1.41 & 1.30 \\
    Age (Images) 	& 11.48 & 11.55 & 11.36 & 11.47 \\
    ISO Speed 		& 1.29 & 1.28 & 1.28 & 1.29 \\
    \hline
  \end{tabular}
\end{table}

\section{Conclusions}\label{sec:conc}
The output of contemporary neural networks, despite being highly accurate in many circumstances, can be considered miscalibrated, thereby producing unreliable output probability estimates~\cite{DBLP:conf/icml/GuoPSW17}. This issue is exacerbated in regression, in which the output of a standard neural network regressor is a point estimate of the predicted values. 

By posing neural network regression as a multi-class classification problem and introducing two novel post-processing calibration methods, we demonstrated that it is possible to produce well-calibrated prediction intervals for neural network regression, that can be critical for a large variety of real-world application. We find that our proposed methods were fast to execute and produce calibration prediction intervals for any desired confidence level, across a variety of regression tasks from the audio and computer vision domains and different neural network architectures. In addition, we found that using a larger number of classification bins generally resulted in tighter prediction intervals, and importantly, that using our proposed methods does not cause any degradation in regression performance, as measured by the mean squared error. 

Future work includes exploring alternative training mechanisms that will lead to tighter calibrated prediction intervals~\cite{keren2017tunable,keren2018fast}, embedding the calibrated outputs into the decision making process of more complex models such as~\cite{keren2018weakly}, and applying the proposed methods to a variety of applications such as computational paralinguistics~\cite{Deng16-EOP,Deng17-REF,Marchi16-TEO}. Further, given the complication when performing regression fusion associated with effects such as multicollinearity, we also plan to test our approach to aid late fusion of multiple neural network regressors.

 \section{Acknowledgements}
 \begin{minipage}{0.15\columnwidth}
 \includegraphics[width=\textwidth]{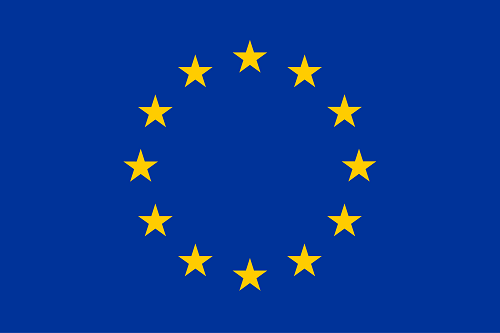}
 \end{minipage}
 \hspace{0.001\columnwidth}
 \begin{minipage}{0.83\columnwidth}
 This work was supported by the European Unions's Seventh Framework and Horizon 2020 Programmes under grant agreements No.\ 338164 (ERC StG iHEARu) and No.\ 688835 (RIA DE-ENIGMA).
 \end{minipage}

\bibliographystyle{IEEEtran}

\bibliography{mybib}

\end{document}